\documentclass{bmvc2k}

\usepackage{cleveref}
\usepackage{filecontents,catchfile}
\usepackage{tabularx}
\usepackage{xcolor}

\title{Few-shot Semantic Segmentation with Self-supervision from Pseudo-classes}

\addauthor{Yiwen Li}{kate@robots.ox.ac.uk}{1}
\addauthor{Gratianus Wesley Putra Data}{gwpd@robots.ox.ac.uk}{1}
\addauthor{Yunguan Fu}{yunguan.fu.18@ucl.ac.uk}{2,3}
\addauthor{Yipeng Hu}{yipeng.hu@ucl.ac.uk}{2}
\addauthor{Victor Adrian Prisacariu}{victor@robots.ox.ac.uk}{1}

\addinstitution{
 Active Vision Lab\\
 University of Oxford\\
 Oxford, UK
}
\addinstitution{
 Centre for Medical Image Computing and Wellcome/EPSRC Centre for Interventional and Surgical Sciences,\\
 University College London,\\
 London, UK
}
\addinstitution{
 InstaDeep,\\
 London, UK
}

\runninghead{Li, Data, Fu, Hu, Prisacariu}{Pseudo-cls Supervised Few-shot Segmentation}

\begin{document}

\maketitle

\begin{abstract}
Despite the success of deep learning methods for semantic segmentation, few-shot semantic segmentation remains a challenging task due to the limited training data and the generalisation requirement for unseen classes.
While recent progress has been particularly encouraging, we discover that existing methods tend to have poor performance in terms of meanIoU when query images contain other semantic classes besides the target class. 
To address this issue, we propose a novel self-supervised task that generates random pseudo-classes in the background of the query images, providing extra training data that would otherwise be unavailable when predicting individual target classes. To that end, we adopted superpixel segmentation for generating the pseudo-classes.
With this extra supervision, we improved the meanIoU performance of the state-of-the-art method by $2.5\%$ and $5.1\%$ on the one-shot tasks, as well as $6.7\%$ and $4.4\%$ on the five-shot tasks, on the PASCAL-5i and COCO benchmarks, respectively.

\end{abstract}

\section{Introduction}
\label{sec:intro}

The goal of semantic segmentation is to classify each pixel in an image into semantically meaningful classes. The data-driven representation learning enabled by deep neural networks has led to accurate and compelling segmentation results~\cite{long2015fully, chen2018encoder}. However, most segmentation methods rely on large datasets with full annotations, which are expensive and labour-intensive to collect. Furthermore, the predictable categories are often constrained to those that have been annotated in the dataset. It is, therefore, of great value to reduce the annotation burden by enabling unseen class prediction with only a few training examples. 
Specifically, given a small number of \emph{support} images with ground truth binary masks of an unseen \emph{target} class, the task is to segment the regions of the same class in unannotated \emph{query} images. 
This is known as few-shot semantic segmentation (FSS)~\cite{shaban2017one}.

While existing methods~\cite{wang2019panet,tian2020prior} have achieved significant improvements over fine-tuning and foreground-background segmentation baselines~\cite{shaban2017one}, we observed a significant drop in performance when the \emph{query} image contains multiple classes
(Figure~\ref{fig:num_class}), \emph{i.e.} the more classes \textcolor{black}{(including both training and testing classes defined in the dataset)} the image has, the more difficult is it to segment the \emph{target} class precisely. 

For example, on the 1-shot PASCAL-5i~\cite{pascal-voc-2012, shaban2017one} benchmark, the current state-of-the-art (PFENet~\cite{tian2020prior}) achieved a meanIoU score of $66.1\%$ for \emph{query} images with one class but only $32.7\%$ for \emph{query} images with four classes. A similar trend has also been observed in the 5-shot results and with other benchmarks such as few-shot COCO~\cite{lin2014microsoft}.

We conjecture that the performance drop is caused by the presence of 
the other non-target classes in the \emph{query} images during training; Here, target class is the class of segmentation interest, whilst in binary segmentation tasks all the non-target classes are
labelled as background regardless of their differences. 
Although the \emph{target}-versus-\emph{non-target} binary classification paradigm seems appropriate for images with single target class, it may inhibit the model from learning meaningful representations from the \emph{non-target} pixels. Given practical limitations such as finite data and imperfect labels, further discriminating these non-target classes can potentially aid classifying \emph{target} pixels, similar to multi-task learning~\cite{he2017mask}. This is especially true in few-shot learning, in which data-efficient model updating is sought-after. 

Following the above intuition, we propose a novel self-supervised training strategy that generates pseudo-classes in the \emph{query} images.
For each episode, a pseudo-class is created by sampling superpixels with high activation from the \emph{query} background. 
Together with the \emph{query} image, an extra pseudo image-mask pair can be generated to assist training.
By guiding the model to discriminate the pseudo-class from the background, this training paradigm enforces the model to distinguish possible non-target classes (those unannotated in the training set) present in the background of the \emph{query} images.
As details presented in the remainder of the paper, the proposed method leads to consistent improvement over different architectures and different datasets.

Our contributions are summarised as follows:
\begin{itemize}
    \setlength\itemsep{-0.2em}
    \item We propose a novel self-supervised task that generates pseudo-classes for few-shot semantic segmentation, which encourage the models to learn a more discriminative feature space during adapting to \emph{novel} target classes.
    \item We highlight the significant performance drop as the number of classes in \emph{query} images increases for existing methods. To the best of our knowledge, it is the first time this issue is highlighted and investigated. 
    \item We present an extensive set of 1-shot and 5-shot experiments that demonstrate the significant improvement from the proposed method, for both PASCAL-5i and COCO datasets. The performance on \emph{query} images with multiple classes were improved and the imbalance between classes were mitigated.
\end{itemize}

\section{Related Work}
\label{sec:related-work}

\subsection{Few-shot Semantic Segmentation}

OSLSM~\cite{shaban2017one} first introduced the task of few-shot segmentation and demonstrated the effectiveness of episodic training, compared favourably with a fine-tuning baseline. PLNet~\cite{dong2018few} extracted class-wise prototypes from the support set and makes predictions based on cosine similarity between pixels and prototypes. This two-step architecture is later inherited by many in this field.

Based on PLNet, extracting information from the support set was proposed to further improve the performance. For instance, PPNet~\cite{liu2020part} and RPMM~\cite{yang2020prototype} generated multiple prototypes for each class through clustering and Expectation-Maximisation. PGNet~\cite{zhang2019pyramid} and DAN~\cite{wang2020few} adopted the attention unit to adjust the foreground prototype according to query pixel features. OANet~\cite{zhao2020objectness} incorporated an objectness segmentation module to calculate the objectness prior. FWB~\cite{Nguyen_2019_ICCV} introduced regularisation to suppress support image background activation and boosts the prototype with an ensemble of experts.

Other improvements can be achieved by refining the feature comparison module. In particular, multi-scale comparison were utilised to overcome spatial inconsistency \cite{zhang2019canet,tian2020prior,zhang2019pyramid,yang2020prototype}. PANet~\cite{wang2019panet} described a prototype alignment regularisation which encouraged the resulted segmentation model to perform few-shot learning in reverse direction.

However, these methods share the common limitation with binary masks, as discussed, during episodic training: pixels not belonging to the sampled target class in query images are all labelled as the same, background class, disregarding how semantically different/similar they are. 
In this work, we address this issue by generating pseudo-classes through superpixels, which creates more training classes and enables discrimination of these non-query classes from the background to provide a better training signal.

\subsection{Superpixel Segmentation}
Superpixel segmentation is an over-segmentation method that groups pixels coherently based on handcrafted features.
It is widely used in image segmentation task to reduce computational costs ~\cite{hwang2019segsort,mivcuvslik2009semantic}.
As superpixel provides additional local information, it has been used for few-shot semantic segmentation to compensate for the lack of annotated data. For example, SSL-ALPNet~\cite{ouyang2020self} used superpixel-based pseudo-labels to replace manual annotation. An important difference in our use of pseudo-label is to amend the performance drop when multiple classes present in the \emph{query} image.
PPNet~\cite{liu2020part} enriched support prototypes by extracting features from unlabelled support images based on superpixel segmentation.
Rather than focusing on the support set, in this work, we sample superpixels to define pseudo-classes only on the background of the query images.

\section{Method}
\label{sec:method}

\begin{figure*}[ht]
    \centering
    \includegraphics[width=.85\linewidth]{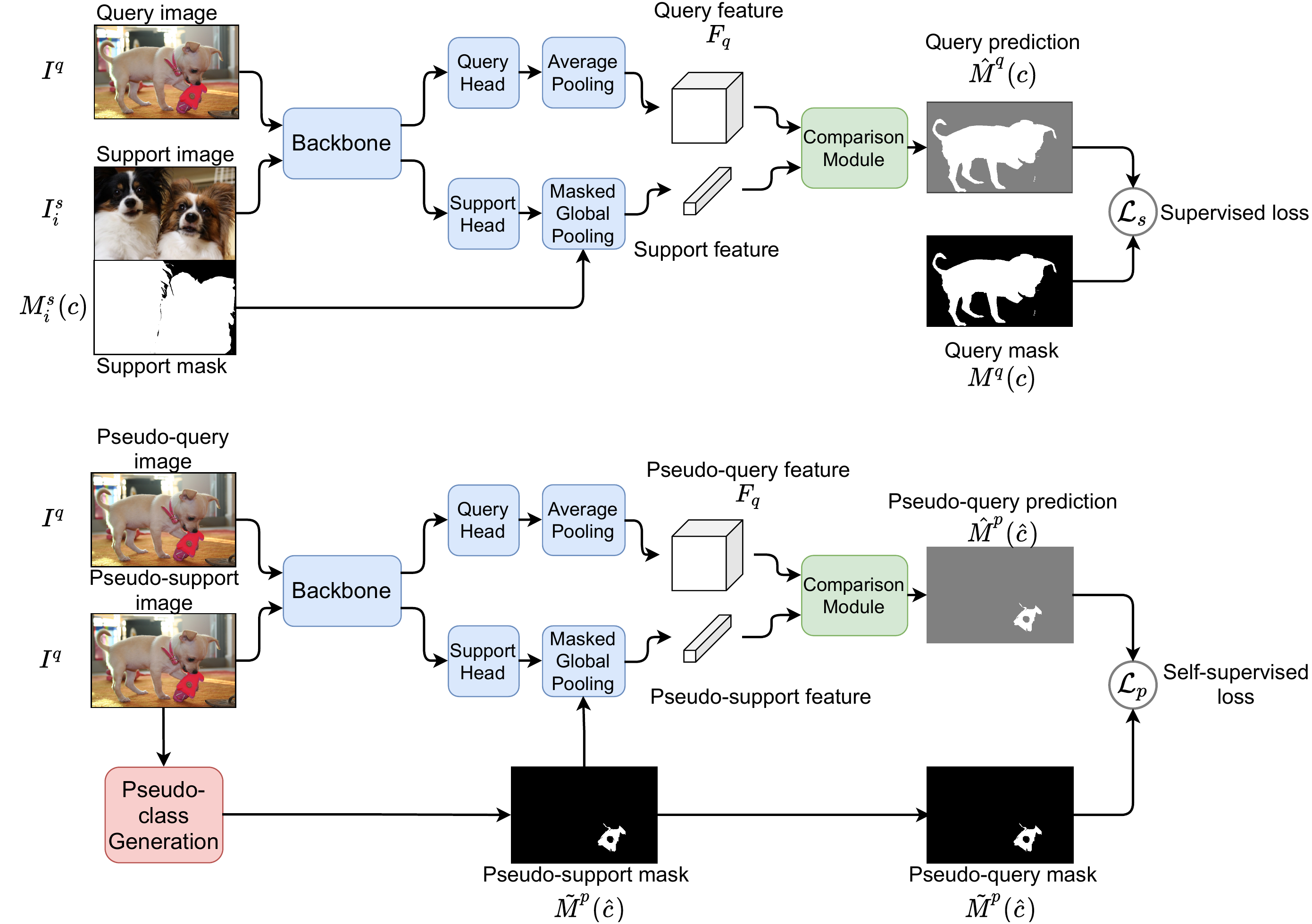}
    \caption{An illustration of training data flow. The upper diagram demonstrates the supervised pathway which predicts the query mask, conditioned on the given support example. The lower diagram demonstrates the self-supervised pathway which generates a pseudo-class and predicts the generated mask, conditioned on the pseudo-support example. Networks are shared between two pathways during training.}
    \label{fig:pipeline}
\end{figure*}
\subsection{Task Description}
\label{sec:task_desc}

Few-shot semantic segmentation aims to learn a model that can segment novel classes when given only a few annotated examples of these classes. 
Specifically, denote $(I, M(c))$ as an image-mask pair, where $I$ represents an RGB image and $M(c)$ represents its binary mask of a class $c$.
The model is trained on a \emph{base} dataset $\mathcal{D}_{base} = \{(I_i, M_i(c)) \mid c \in \mathcal{C}_{base}\}^N_{i=1}$ and tested on a \emph{novel} dataset $\mathcal{D}_{novel} = \{(I_j, M_j(c)) \mid c \in \mathcal{C}_{novel}\}^{N'}_{j=1}$, where \emph{base} classes and \emph{novel} classes are mutually exclusive \emph{i.e.} $\mathcal{C}^\text{base} \cap \mathcal{C}^\text{novel} = \emptyset$.

Following the benchmark introduced by \cite{shaban2017one}, the evaluation takes place in an episodic manner: a \emph{target} class $c \in C_{novel}$ is sampled for each \emph{episode}, consisting of a \emph{support} set $\mathcal{S} = \{ (I^s_i, M^s_i(c)) \}^k_{i=1}$ and a \emph{query} image-mask pair $(I^q, M^q(c))$ of the \emph{target} class. The model is tasked to predict the \emph{query} mask $M^q(c)$ given the \emph{query} image $I_q$ and the \emph{support} set $\mathcal{S}$.  

\begin{figure}
\centering
\begin{tabular}{cccc}
\includegraphics[width=2.5cm]{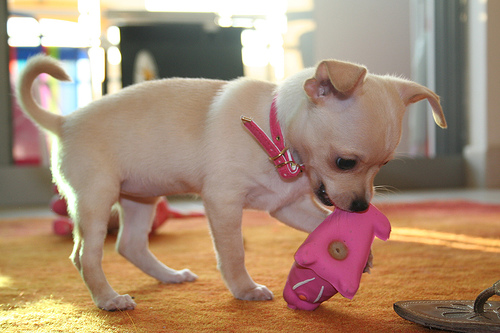} &
\includegraphics[width=2.5cm]{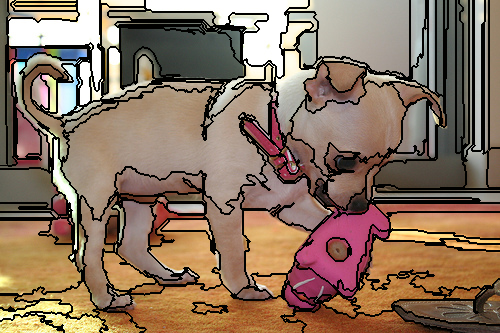} &
\includegraphics[width=2.5cm]{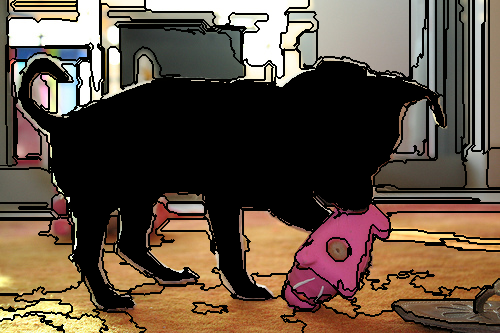} &
\includegraphics[width=2.5cm]{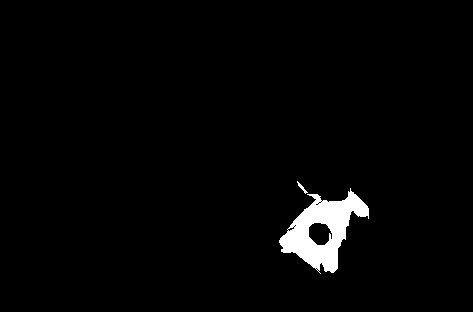}
\\
(a)&(b)&(c)&(d)
\end{tabular}
\caption{Step-by-step breakdown of pseudo-mask generation for self-supervision. (a) Input query image. (b) Generate superpixels in the query image. (c) Remove superpixels that coincide with the original class ground truth. (d) Randomly sample superpixel with high corresponding feature activation, and convert it into a binary mask.}
\label{fig:superpixel_gen}
\end{figure}

\subsection{Self-supervision from Pseudo-classes}
\label{sec:superpixel_method}

Most state-of-the-art few-shot semantic segmentation methods follow the episodic training paradigm.
For each iteration, an \emph{episode}, consisting a \emph{support} set $\mathcal{S} = \{ (I^s_i, M^s_i(c)) \}^k_{i=1}$ and a \emph{query} image-mask pair $(I^q, M^q(c))$, where the \emph{target} class $c \in C_{base}$ is sampled from the \emph{base} dataset $\mathcal{D}_{base}$.

As shown in Figure~\ref{fig:pipeline} both \emph{query} and \emph{support} images are encoded by the feature extractor into feature maps. The \emph{target} class prototype is then derived through global average pooling all foreground pixels of the support feature maps. The comparison module makes the final prediction based on the input \emph{query} feature maps and \emph{target} class prototype. 

The model is optimised to reduce $\mathcal{L}(M^q(c),\hat{M}^q(c))$, the cross entropy loss between the predicted \emph{query} mask $\hat{M}^q(c)$ and the ground-truth \emph{query} mask $M^q(c)$.
Denote $\mathcal{L}(M,\hat{M})$ as the spatially averaged cross-entropy loss between ground-truth $M$ and prediction $\hat{M}$:
\begin{equation}
     \mathcal{L}(M,\hat{M}) = -\frac{1}{WH} \sum^{W}_{x=1} \sum^{H}_{y=1} \left[ M_{(x,y)} \log \hat{M}_{(x,y)} + (1 - M_{(x,y)}) \log (1 - \hat{M}_{(x,y)})\right],
\end{equation}
where $M_{(x,y)}$ represents the value of mask $M$ at pixel $(x,y)$.

Under this training paradigm, \emph{non-target}-class objects with different semantic meanings in the \emph{query} image are all regarded as the same class -- background. The purposed self-supervision task aims to better utilise the information from the background pixels.

First, we apply a superpixel segmentation method (such as Felzenszwalb \emph{et al.}~\cite{felzenszwalb2004efficient}) to $I^q$. For each superpixel, denote its region as the pseudo-class $\Dot{c}$ and the binary mask representing the region as the pseudo-mask $M^p(\Dot{c})\in\{0,1\}^{H\times W}$.
Second, we refine each pseudo-mask $M^p(\Dot{c})$ with the aid of the ground-truth mask $M^q(c)$ by removing pixels of the \emph{target} class $c$,
\begin{align}
    \tilde{M}^p(\Dot{c}) = M^p(\Dot{c}) \odot (1 - M^q(c)),
\end{align}
where $\odot$ denote the element-wise multiplication.
Then, a class activation score $s(\Dot{c})$ is calculated for each pseudo-class $\Dot{c}$ by averaging the extracted query feature $F^q$ over the pseudo-mask foreground and channels, specifically:
\begin{align}
    s(\Dot{c}) &= \frac{\sum^{W}_{x=1} \sum^{H}_{y=1} \sum^{d}_{z=1} F^q_{(x,y,z)} \tilde{M}^p_{(x,y)}(\Dot{c})}{\sum^{W}_{x=1} \sum^{H}_{y=1} \sum^{d}_{z=1} \tilde{M}^p_{(x,y)}(\Dot{c})}.
    \label{eq:activation_score}
\end{align}

Lastly, a pseudo-class $\hat{c}$ is randomly selected among the ones having the top-5 scores, so that the selected pseudo-class is more likely to be non-background while keeping diversity between pseudo-classes.
The corresponding refined mask is denoted as $\tilde{M}^p(\hat{c})$.

As shown in Figure~\ref{fig:pipeline}, the additional image-mask pair $(I^q, \tilde{M}^p(\hat{c}))$ serves as both \emph{pseudo-support} and \emph{pseudo-query}, forming a \emph{pseudo support-query} pair for $I^q$ in the current training iteration. Specifically, the self-supervised task is defined as: given the \emph{pseudo-support} example $(I^q, \tilde{M}^p(\hat{c}))$, the model is required to predict the same mask $\tilde{M}^p(\hat{c})$ for the pseudo-class $\hat{c}$. The corresponding prediction is denoted as $\hat{M}^p(\hat{c})\in [0, 1]^{H \times W}$.
Similarly, we compute a cross-entropy loss $\mathcal{L}(\tilde{M}^p(\hat{c}), \hat{M}^p(\hat{c})),$ for this self-supervision.

The total loss is therefore a weighted sum of the supervised loss with the target class $c$ and the self-supervised loss with the pseudo-class $\hat{c}$,
\begin{equation}\label{eq:total_loss}
    \mathcal{L}_\text{total} = \mathcal{L}(M^q(c), \hat{M}^q(c)) + \alpha \mathcal{L}(\tilde{M}^p(\hat{c}), \hat{M}^p(\hat{c})),
\end{equation}
where $\alpha$ is a scaling coefficient to control the contribution of the self-supervised loss towards the overall objective. 

\section{Experiments}

\subsection{Implementation Details}
\label{sec:imp_details}

We evaluated our methods on the PASCAL-5i~\cite{shaban2017one} (which consists of PASCAL VOC 2012~\cite{pascal-voc-2012} with extended annotations from SDS~\cite{hariharan2014simultaneous}) and COCO~\cite{lin2014microsoft} datasets. Each dataset has 4 configurations (folds) of base-novel class splits, following OSLSM~\cite{shaban2017one} and FWB~\cite{Nguyen_2019_ICCV} protocol.
We chose PFENet~\cite{tian2020prior}, the current state-of-the-art, as our baseline architecture. 
We also followed the training and evaluation setting of the original paper~\cite{tian2020prior} while removing the train-free prior generation module. Felzenszwalb \emph{et al.}~\cite{felzenszwalb2004efficient} was used for superpixel segmentation with scale set to 100, sigma to 0.8 and min\_size to 200. The derived superpixel segmetations were further refined by the training (base) class masks. The hyper-parameter $\alpha$ was set to $0.5$ in equation \eqref{eq:total_loss}.
We report the meanIoU metric for evaluation, which is the intersection-over-union (IoU) averaged over classes.
For 5-shot evaluation, the individual 1-shot models were used without fine-tuning and we averaged the resulting prototype vectors of supports before feeding them to the comparison module. For brevity, the prefix ``SS-'' denotes models trained with the proposed self-supervision.

\subsection{Results}
\label{sec:result}

\newcommand{\pascaltable}{\small
\def\tabcolsep{0.8pt}
\begin{tabular}{|c|ccccc|ccccc|}
    \hline
                                        & \multicolumn{5}{c|}{1-shot}                                                  & \multicolumn{5}{c|}{5-shot} \\
    method                              & fold1                 & fold2                 & fold3                 & fold4                 & mean                  & fold1                         & fold2                & fold3                 & fold4                 & mean \\
    \hline
    \multicolumn{11}{|c|}{VGG-16}       \\\hline
    OSLSM~\cite{shaban2017one}          & 33.6                  & 55.3                  & 40.9                  & 33.5                  & 40.8                  & 37.5                          & 50.0                 & 44.1                  & 33.9                  & 41.4 \\
  co\_FCN~\cite{rakelly2018conditional} & 36.7                  & 50.6                  & 44.9                  & 32.4                  & 41.2                  & 35.9                            & 58.1                 & 42.7                  & 39.1                  & 44.0 \\
    AMP~\cite{siam2019amp}              & 41.9                  & 50.2                  & 46.7                  & 34.7                  & 43.4                  & 41.8                          & 55.5                 & 50.3                  & 39.9                  & 46.9 \\
    Sg-one~\cite{zhang2020sg}           & 40.2                  & 58.4                  & 48.4                  & 38.4                  & 46.4                  & 41.9                          & 58.6                 & 48.6                  & 39.4                  & 47.1 \\
    FWBNet~\cite{Nguyen_2019_ICCV}      & 47.0                  & 59.6                  & 52.6                  & 48.3                  & 51.9                  & 50.9                          & 62.9                 & 56.5                  & 50.1                  & 55.1 \\
    RPPM~\cite{yang2020prototype}       & 47.1                  & 65.8                  & 50.6                  & 48.5                  & 53.0                  & 50.0                          & 66.5                 & 51.9                  & 47.6                  & 54.0 \\    
    CRNet~\cite{liu2020crnet}           & -                     & -                     & -                     & -                     & 55.2                 & -                             &                      & -                     & -                     & 58.5 \\
    \hline
    PANet~\cite{wang2019panet}          & 42.3                  & 58.0                  & 51.1                  & 41.3                  & 48.2                  & 51.8                          & \textbf{64.6}        & 59.8                  & 46.5                  & 55.7 \\
    SS-PANet                            & \textbf{49.3}         & \textbf{60.8}         & \textbf{53.9}         & \textbf{45.2}         & \textbf{52.3}         & \textbf{54.5}                 & 64.5                 & \textbf{62.3}         & \textbf{52.0}         & \textbf{58.3} \\\hline
    PFENet\cite{tian2020prior}          & \textbf{56.9}         & \textbf{68.2}         & 54.4                  & 52.4                  & 58.0                  & \textbf{59.0}                 & 69.1                 & 54.8                  & 52.9                  & 59.0 \\
    SS-PFENet                           & 54.5                  & 67.4                  & \textbf{63.4}         & \textbf{54.0}         & \textbf{59.8}         & 56.9                          & \textbf{70.0}        & \textbf{68.3}         & \textbf{62.1}         & \textbf{64.3} \\
    \hline
    \multicolumn{11}{|c|}{ResNet-50}    \\\hline
    PPNet~\cite{liu2020part}            & 48.6                  & 60.6                  & 55.7                  & 46.5                  & 52.9                  & 58.9                          & 68.3                 & 66.8                  & 58.0                  & 63.0 \\
    RPMM~\cite{yang2020prototype}       & 50.2                  & 66.9                  & 52.6                  & 50.7                  & 55.1                  & 56.3                          & 68.3                 & 54.5                  & 51.0                  & 57.5 \\
    CANet~\cite{zhang2019canet}         & 52.5                  & 65.9                  & 51.3                  & 51.9                  & 55.4                  & 55.5                          & 67.8                 & 51.9                  & 53.2                  & 57.1 \\
    PGNet~\cite{zhang2019pyramid}       & 56.0                  & 66.9                  & 50.6                  & 50.4                  & 56.0                  & 57.7                          & 68.7                 & 52.9                  & 54.6                  & 58.5 \\
    ORG~\cite{Ying_2021_WACV}           & 52.6                  & 65.8                  & 54.7                  & 52.1                  & 56.3                  & 57.2                          & 67.8                 & 57.5                  & 56.2                  & 59.7 \\
    OANet~\cite{zhao2020objectness}     & 56.9                  & 66.4                  & 57.1                  & 50.7                  & 57.8                  & 59.6                          & 68.9                 & 57.8                  & 54.4                  & 60.2 \\
    ASGNet~\cite{li2021adaptive}        & 58.8                  & 67.9                  & 56.8                  & 53.7                  & 59.3                  & 63.66                         & 70.6                 & 64.2                  & 57.4                  & 64.0 \\
    RePRI~\cite{boudiaf2021few}         & 60.2                  & 67.0                  & 61.7                  & 47.5                  & 59.1                  & \textbf{$64.5^*$}             & 70.8                 & \textbf{$71.7^*$}     & 60.3                  & 66.8 \\
    SCL~\cite{zhang2021self}            & \textbf{$63.0^*$}        & \textbf{$70.0^*$} & 56.5                   & \textbf{$57.7^*$}     & 61.8                  & \textbf{$64.5^*$}             & 70.9                 & 57.3                  & 58.7                  & 62.9 \\
    \hline
    PFENet~\cite{tian2020prior}         & 61.7                  & 69.5                  & 55.4                  & 56.3                  & 60.7                  & 63.1                          & 70.7                 & 55.8                  & 57.9                  & 61.9 \\
    
    SS-PFENet                           & 58.9                  & 69.9                  & \textbf{$66.4^*$}     & \textbf{$57.7^*$}     & \textbf{$63.2^*$}      & 61.4                         & \textbf{$75.0^*$}    & 70.5                  & \textbf{$67.7^*$}     & \textbf{$68.6^*$}   \\
    \hline
\end{tabular}}
\begin{table}[ht]
    \centering
    \small
\def\tabcolsep{0.8pt}
\begin{tabular}{|c|ccccc|ccccc|}
    \hline
                                        & \multicolumn{5}{c|}{1-shot}                                                  & \multicolumn{5}{c|}{5-shot} \\
    method                              & fold1                 & fold2                 & fold3                 & fold4                 & mean                  & fold1                         & fold2                & fold3                 & fold4                 & mean \\
    \hline
    \multicolumn{11}{|c|}{VGG-16}       \\\hline
    OSLSM~\cite{shaban2017one}          & 33.6                  & 55.3                  & 40.9                  & 33.5                  & 40.8                  & 37.5                          & 50.0                 & 44.1                  & 33.9                  & 41.4 \\
  co\_FCN~\cite{rakelly2018conditional} & 36.7                  & 50.6                  & 44.9                  & 32.4                  & 41.2                  & 35.9                            & 58.1                 & 42.7                  & 39.1                  & 44.0 \\
    AMP~\cite{siam2019amp}              & 41.9                  & 50.2                  & 46.7                  & 34.7                  & 43.4                  & 41.8                          & 55.5                 & 50.3                  & 39.9                  & 46.9 \\
    Sg-one~\cite{zhang2020sg}           & 40.2                  & 58.4                  & 48.4                  & 38.4                  & 46.4                  & 41.9                          & 58.6                 & 48.6                  & 39.4                  & 47.1 \\
    FWBNet~\cite{Nguyen_2019_ICCV}      & 47.0                  & 59.6                  & 52.6                  & 48.3                  & 51.9                  & 50.9                          & 62.9                 & 56.5                  & 50.1                  & 55.1 \\
    RPPM~\cite{yang2020prototype}       & 47.1                  & 65.8                  & 50.6                  & 48.5                  & 53.0                  & 50.0                          & 66.5                 & 51.9                  & 47.6                  & 54.0 \\    
    CRNet~\cite{liu2020crnet}           & -                     & -                     & -                     & -                     & 55.2                 & -                             &                      & -                     & -                     & 58.5 \\
    \hline
    PANet~\cite{wang2019panet}          & 42.3                  & 58.0                  & 51.1                  & 41.3                  & 48.2                  & 51.8                          & \textbf{64.6}        & 59.8                  & 46.5                  & 55.7 \\
    SS-PANet                            & \textbf{49.3}         & \textbf{60.8}         & \textbf{53.9}         & \textbf{45.2}         & \textbf{52.3}         & \textbf{54.5}                 & 64.5                 & \textbf{62.3}         & \textbf{52.0}         & \textbf{58.3} \\\hline
    PFENet\cite{tian2020prior}          & \textbf{56.9}         & \textbf{68.2}         & 54.4                  & 52.4                  & 58.0                  & \textbf{59.0}                 & 69.1                 & 54.8                  & 52.9                  & 59.0 \\
    SS-PFENet                           & 54.5                  & 67.4                  & \textbf{63.4}         & \textbf{54.0}         & \textbf{59.8}         & 56.9                          & \textbf{70.0}        & \textbf{68.3}         & \textbf{62.1}         & \textbf{64.3} \\
    \hline
    \multicolumn{11}{|c|}{ResNet-50}    \\\hline
    PPNet~\cite{liu2020part}            & 48.6                  & 60.6                  & 55.7                  & 46.5                  & 52.9                  & 58.9                          & 68.3                 & 66.8                  & 58.0                  & 63.0 \\
    RPMM~\cite{yang2020prototype}       & 50.2                  & 66.9                  & 52.6                  & 50.7                  & 55.1                  & 56.3                          & 68.3                 & 54.5                  & 51.0                  & 57.5 \\
    CANet~\cite{zhang2019canet}         & 52.5                  & 65.9                  & 51.3                  & 51.9                  & 55.4                  & 55.5                          & 67.8                 & 51.9                  & 53.2                  & 57.1 \\
    PGNet~\cite{zhang2019pyramid}       & 56.0                  & 66.9                  & 50.6                  & 50.4                  & 56.0                  & 57.7                          & 68.7                 & 52.9                  & 54.6                  & 58.5 \\
    ORG~\cite{Ying_2021_WACV}           & 52.6                  & 65.8                  & 54.7                  & 52.1                  & 56.3                  & 57.2                          & 67.8                 & 57.5                  & 56.2                  & 59.7 \\
    OANet~\cite{zhao2020objectness}     & 56.9                  & 66.4                  & 57.1                  & 50.7                  & 57.8                  & 59.6                          & 68.9                 & 57.8                  & 54.4                  & 60.2 \\
    ASGNet~\cite{li2021adaptive}        & 58.8                  & 67.9                  & 56.8                  & 53.7                  & 59.3                  & 63.66                         & 70.6                 & 64.2                  & 57.4                  & 64.0 \\
    RePRI~\cite{boudiaf2021few}         & 60.2                  & 67.0                  & 61.7                  & 47.5                  & 59.1                  & \textbf{$64.5^*$}             & 70.8                 & \textbf{$71.7^*$}     & 60.3                  & 66.8 \\
    SCL~\cite{zhang2021self}            & \textbf{$63.0^*$}        & \textbf{$70.0^*$} & 56.5                   & \textbf{$57.7^*$}     & 61.8                  & \textbf{$64.5^*$}             & 70.9                 & 57.3                  & 58.7                  & 62.9 \\
    \hline
    PFENet~\cite{tian2020prior}         & 61.7                  & 69.5                  & 55.4                  & 56.3                  & 60.7                  & 63.1                          & 70.7                 & 55.8                  & 57.9                  & 61.9 \\
    
    SS-PFENet                           & 58.9                  & 69.9                  & \textbf{$66.4^*$}     & \textbf{$57.7^*$}     & \textbf{$63.2^*$}      & 61.4                         & \textbf{$75.0^*$}    & 70.5                  & \textbf{$67.7^*$}     & \textbf{$68.6^*$}   \\
    \hline
\end{tabular}
    \caption{meanIoU results on PASCAL-5i~\cite{shaban2017one}. Best performance with the same model and overall are respectively bolded and starred. ``SS-'' refers to models with self-supervision.}
    \label{tab:pascal_table}
\end{table}

\newcommand{\cocotable}{\small
\def\tabcolsep{0.8pt}
\begin{tabular}{|c|ccccc|ccccc|}
    \hline
                                    & \multicolumn{5}{c|}{1-shot}                                                  & \multicolumn{5}{c|}{5-shot}  \\
    method                          & fold1                 & fold2                 & fold3                 & fold4                 & mean                  & fold1                             & fold2                & fold3         & fold4         & mean \\
    \hline
    \multicolumn{11}{|c|}{VGG-16}   \\\hline
    FWBNet~\cite{Nguyen_2019_ICCV}  & 18.4                  & 16.7                  & 19.6                  & 25.4                  & 20.0                  & 20.9                              & 19.2                 & 21.9                  & 28.4                  & 22.6 \\
    \hline
    PANet~\cite{wang2019panet}      & 28.7                  & 21.2                  & 19.1                  & 14.8                  & 21.0                  & 39.4                              & 28.3                 & 28.2                  & 22.7                  & 29.7 \\
    SS-PANet                        & \textbf{29.8}         & \textbf{21.2}         & \textbf{26.5}         & \textbf{28.5}         & \textbf{26.2}         & \textbf{36.7}                     & \textbf{41.0}        & \textbf{37.6}         & \textbf{35.6}         & \textbf{37.7} \\\hline
    PFENet~\cite{tian2020prior}     & 34.3                  & 33.0                  & 32.3                  & 30.1                  & 32.4                  & 38.5                              & 38.6                 & 38.2                  & 34.3          & 37.4 \\
    SS-PFENet                       &\textbf{$35.6^*$}      &\textbf{$39.2^*$}      &\textbf{$37.6^*$}      &\textbf{$37.3^*$}      &\textbf{$37.5^*$}      &\textbf{$40.4^*$} &\textbf{$45.8^*$}     &\textbf{$40.3^*$}      &\textbf{$40.7^*$}      & \textbf{$41.8^*$} \\
    \hline

\end{tabular}}
\begin{table}[ht]
    \centering
    \small
\def\tabcolsep{0.8pt}
\begin{tabular}{|c|ccccc|ccccc|}
    \hline
                                    & \multicolumn{5}{c|}{1-shot}                                                  & \multicolumn{5}{c|}{5-shot}  \\
    method                          & fold1                 & fold2                 & fold3                 & fold4                 & mean                  & fold1                             & fold2                & fold3         & fold4         & mean \\
    \hline
    \multicolumn{11}{|c|}{VGG-16}   \\\hline
    FWBNet~\cite{Nguyen_2019_ICCV}  & 18.4                  & 16.7                  & 19.6                  & 25.4                  & 20.0                  & 20.9                              & 19.2                 & 21.9                  & 28.4                  & 22.6 \\
    \hline
    PANet~\cite{wang2019panet}      & 28.7                  & 21.2                  & 19.1                  & 14.8                  & 21.0                  & 39.4                              & 28.3                 & 28.2                  & 22.7                  & 29.7 \\
    SS-PANet                        & \textbf{29.8}         & \textbf{21.2}         & \textbf{26.5}         & \textbf{28.5}         & \textbf{26.2}         & \textbf{36.7}                     & \textbf{41.0}        & \textbf{37.6}         & \textbf{35.6}         & \textbf{37.7} \\\hline
    PFENet~\cite{tian2020prior}     & 34.3                  & 33.0                  & 32.3                  & 30.1                  & 32.4                  & 38.5                              & 38.6                 & 38.2                  & 34.3          & 37.4 \\
    SS-PFENet                       &\textbf{$35.6^*$}      &\textbf{$39.2^*$}      &\textbf{$37.6^*$}      &\textbf{$37.3^*$}      &\textbf{$37.5^*$}      &\textbf{$40.4^*$} &\textbf{$45.8^*$}     &\textbf{$40.3^*$}      &\textbf{$40.7^*$}      & \textbf{$41.8^*$} \\
    \hline

\end{tabular}
    \caption{meanIoU results on COCO~\cite{pascal-voc-2012}. Best performance with the same model and overall are respectively bolded and starred. ``SS-'' refers to models with self-supervision.}
    \label{tab:coco_table}
\end{table}

The meanIoU performances per fold for PASCAL-5i 1-shot and 5-shot tasks are presented in Table~\ref{tab:pascal_table}.
By incorporating the self-supervision during the training of ResNet-50 PFENet, we achieved $63.2\%$ meanIoU on the PASCAL-5i 1-shot task, improving its baseline performance by $2.5\%$. For the 5-shot task, we achieved $68.6\%$ meanIoU with a $6.7\%$ absolute improvement. This sets a new state-of-the-art on the PASCAL-5i dataset for both 1-shot and 5-shot.
On the COCO benchmark (Table~\ref{tab:coco_table}), we achieved $37.5\%$ and $41.8\%$ meanIoU for 1-shot and 5-shot using VGG-16 PFENet and set the new state-of-the-art from our knowledge.

\begin{figure}
\begin{tabular}{cc}
\includegraphics[width=3.9cm]{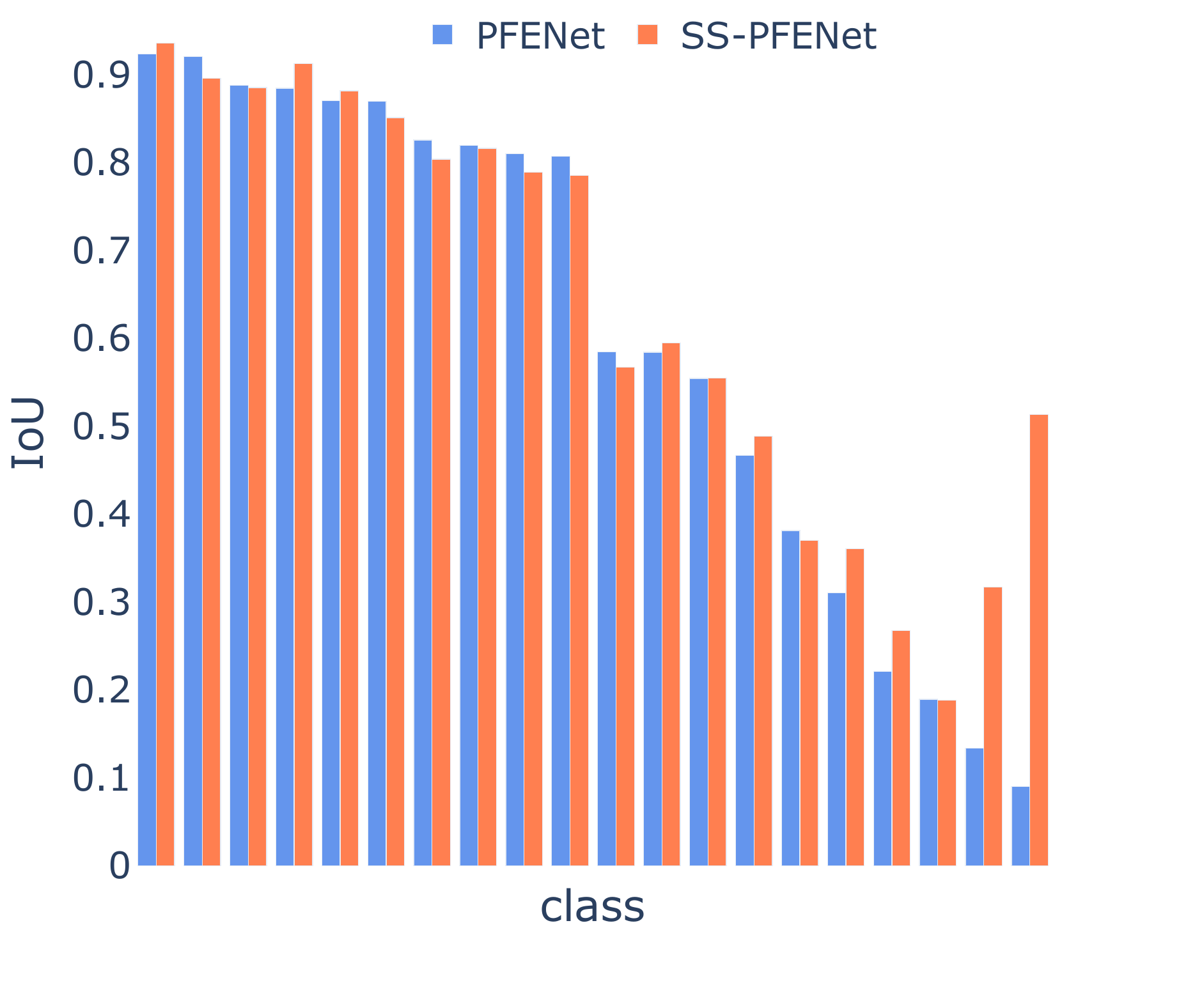} &
\includegraphics[width=7.4cm]{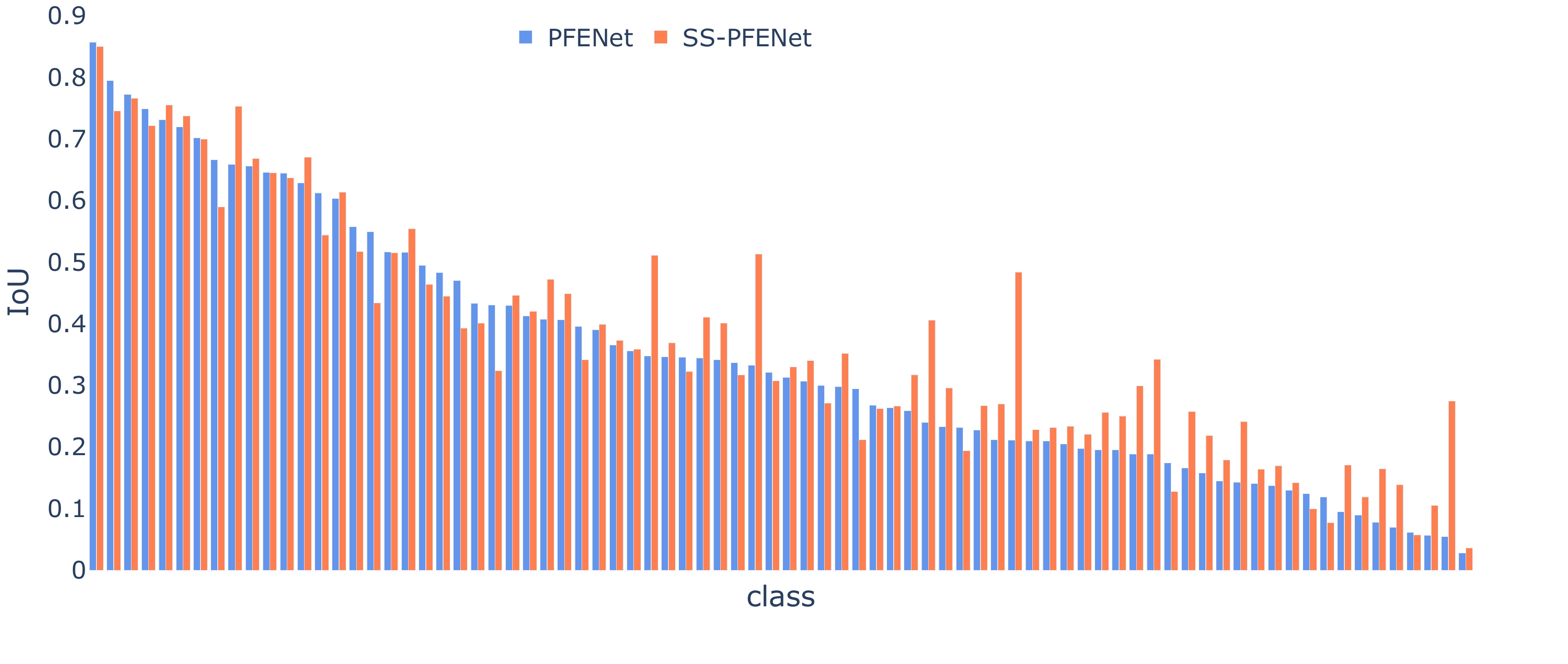}
\\
(a) PASCAL-5i & (b) COCO
\end{tabular}
\caption{Class IoUs reached by PFENet with and without superpixel self-supervision. X-axis corresponds to different classes. A larger version with class names annotated is available in the supplementry material.}
\label{fig:iou_by_class}
\end{figure}

Additionally, we present the IoU values per class in Figure~\ref{fig:iou_by_class} for PFENet with and without self-supervision on both PASCAL-5i and COCO. Most of the improvement came from previously under-performing classes (i.e person, dining table, potted plant, etc.), and thus led to a more class-wise balanced performance. 

\begin{figure}[ht]
\centering
\begin{tabular}{ccc}
\includegraphics[width=2.8cm]{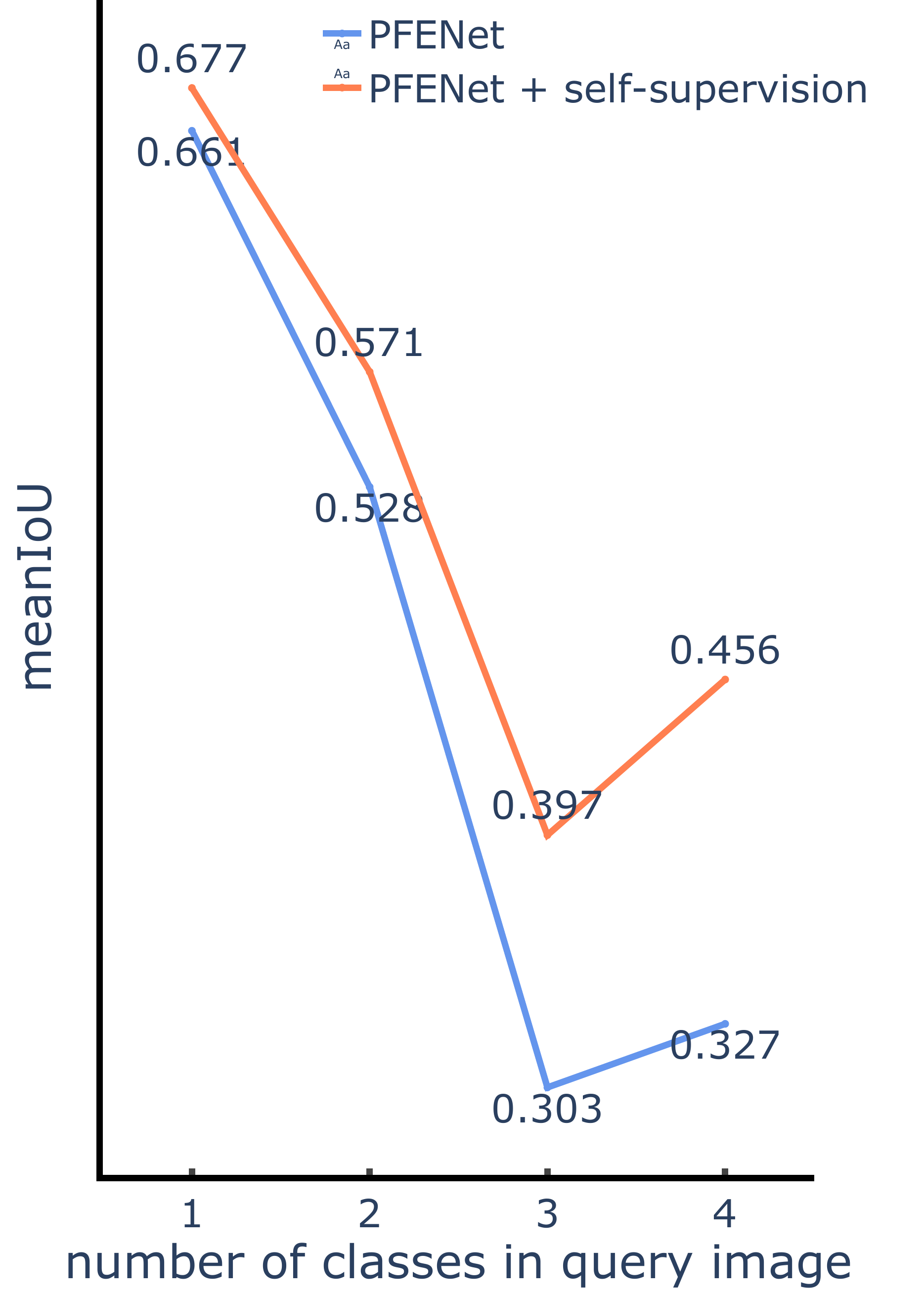} &
\includegraphics[width=2.8cm]{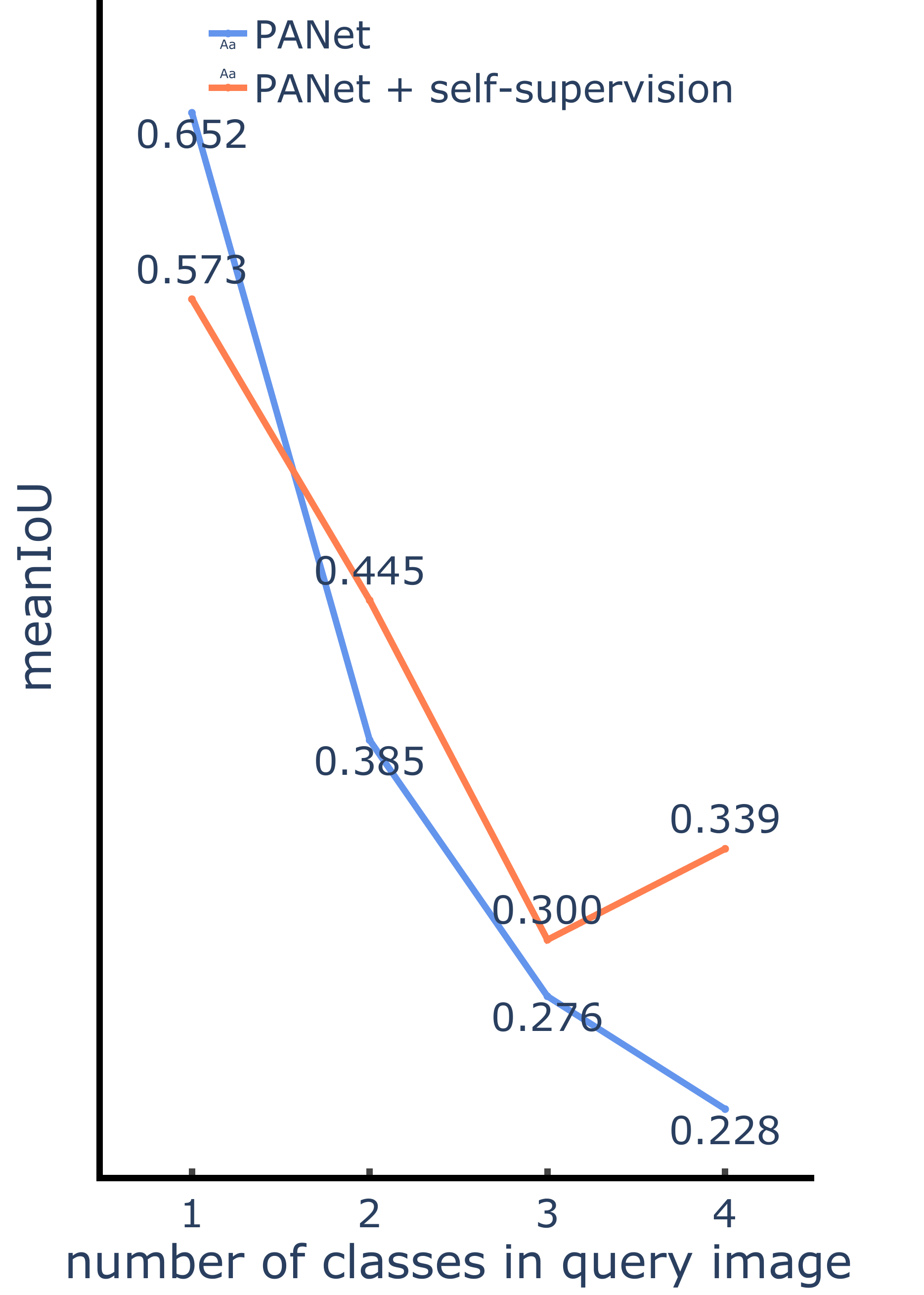} &
\includegraphics[width=5.6cm]{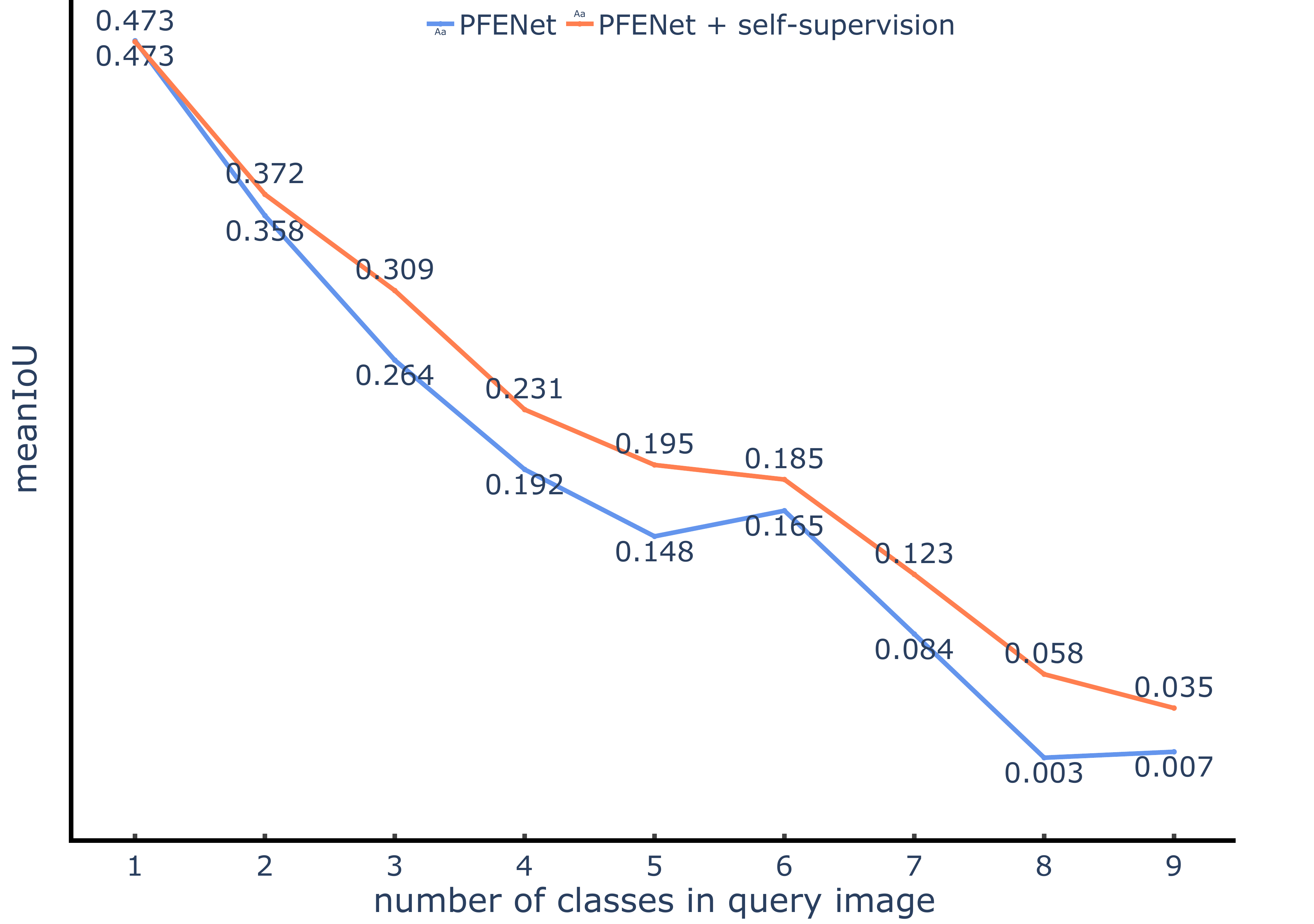} \\
(a) PASCAL-5i & (b) PASCAL-5i & (c) COCO
\end{tabular}
\caption{meanIoUs reached with and without superpixel self-supervision when different number of classes occur in the query image.}
\label{fig:num_class}
\end{figure}

The meanIoU performances on images with the different number of classes are plotted in Figure~\ref{fig:num_class}. \textcolor{black}{Here the number of classes is defined by the ground-truth labels. Therefore, for PASCAL and COCO, the maximum available numbers of classes in a query image are respectively 20 and 80.} From Figure~\ref{fig:num_class}(a) and Figure~\ref{fig:num_class}(c), one can observe that the proposed method consistently improved the performance when multiple classes were presented in the query image over both PASCAL-5i and COCO dataset. This explains the higher relative improvement in performance achieved on COCO ($11.8\%$) compare to PASCAL-5i ($3.1\%$) when the same backbone (VGG-16) was adopted.

\begin{figure}
    \centering
    \includegraphics[width=8.0cm]{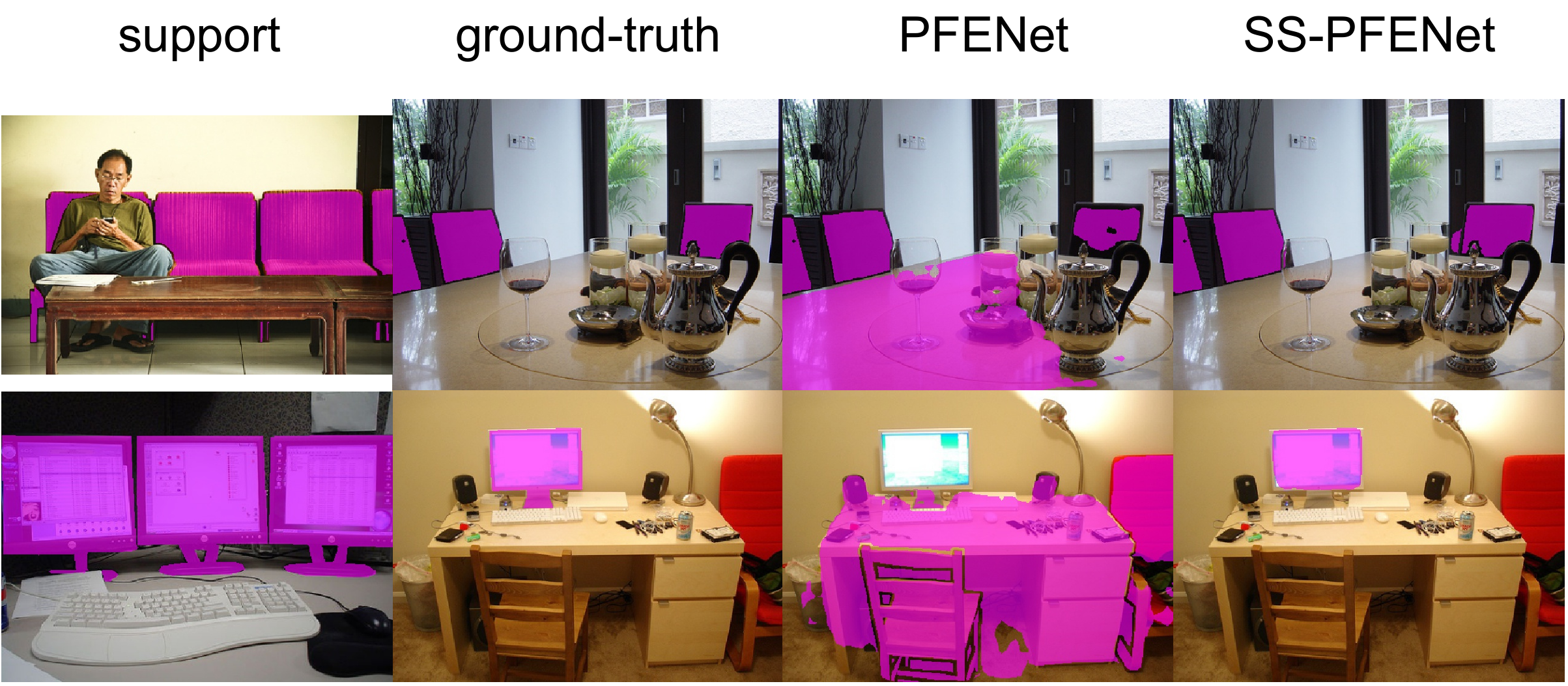}
    \caption{Qualitative results of ResNet-50 PFENet with and without the purposed supervision. More examples are available in the supplementary materials.}
    \label{fig:qualitative_example}
\end{figure}

\textcolor{black}{
Figure~\ref{fig:qualitative_example} shows some qualitative examples achieved by ResNet-50 PFENet with and without the purposed self-supervision. When multiple objects exist in the query image, PFENet succeeds in detecting them from the background but fails to filter out non-target-class objects. This failure in discriminating class differences has been rectified by the proposed self-supervised task which is likely to increase class diversities due to introducing pseudo-classes.
}

\begin{figure}[ht]
\begin{tabular}{cccc}
\includegraphics[width=2.8cm]{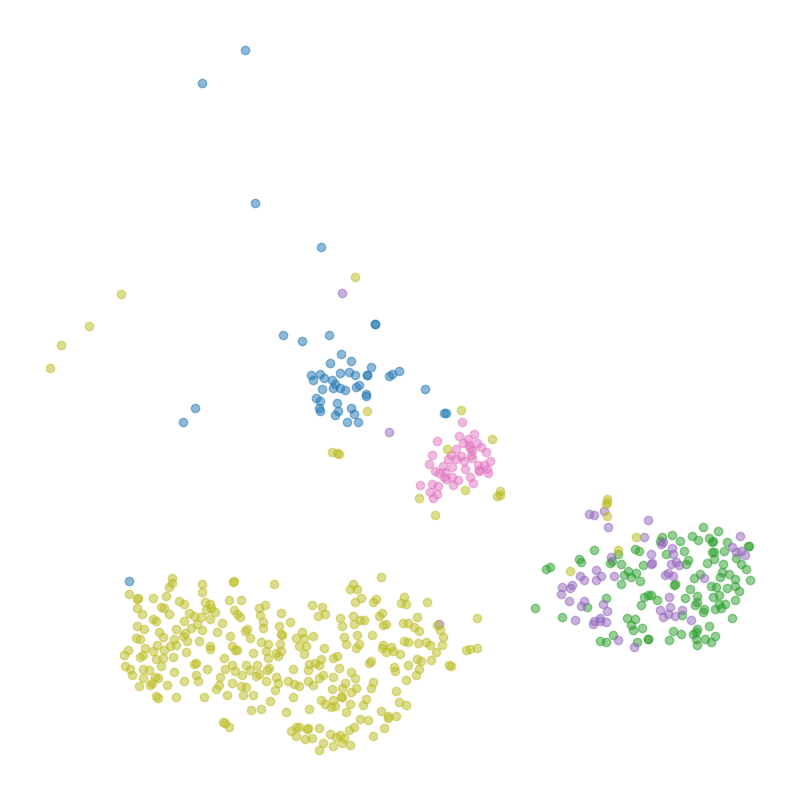} &
\includegraphics[width=2.8cm]{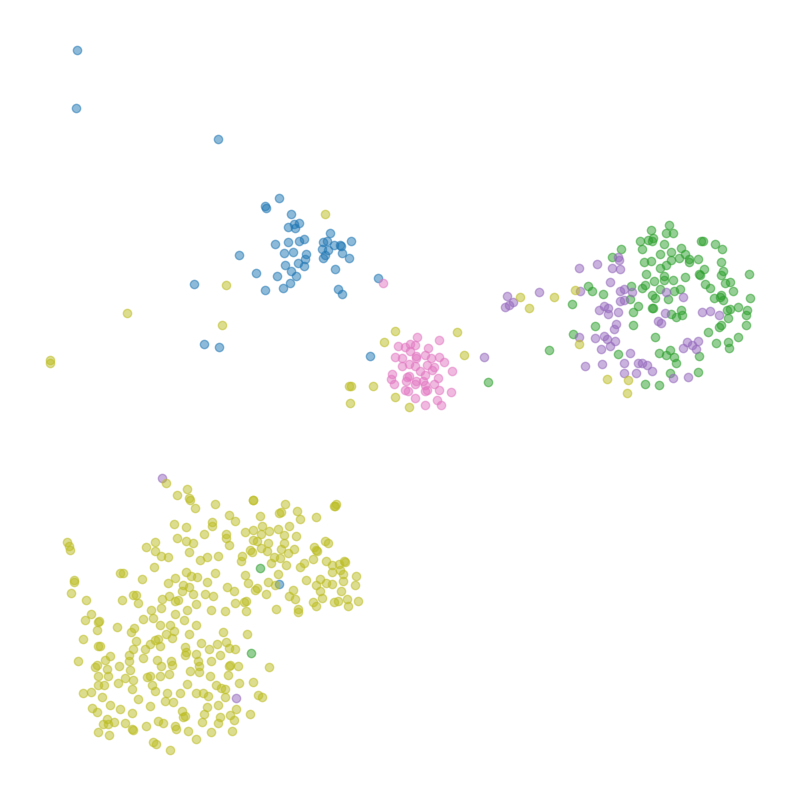} &
\includegraphics[width=2.8cm]{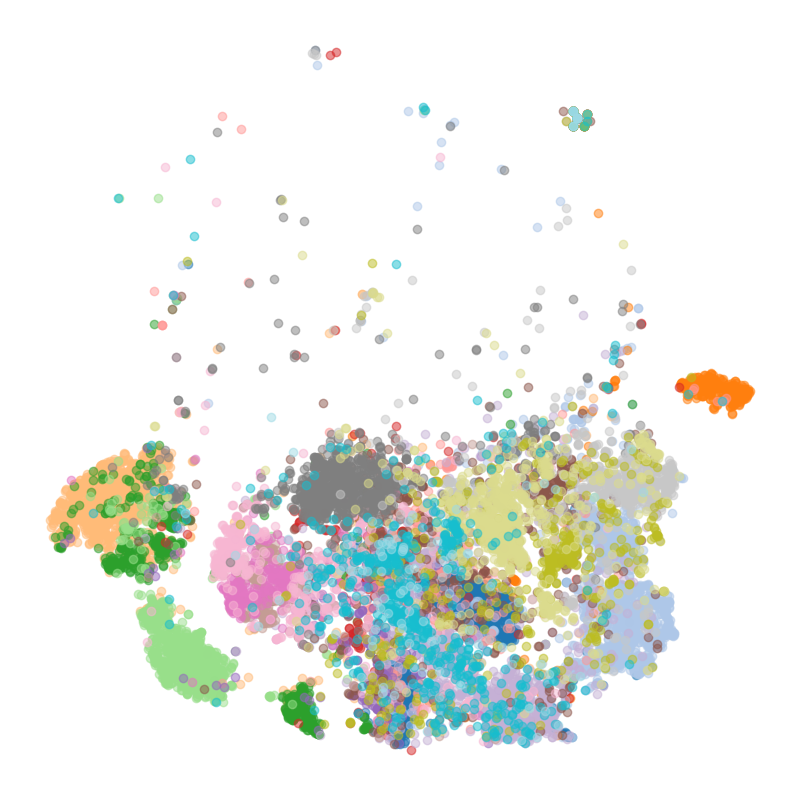} &
\includegraphics[width=2.8cm]{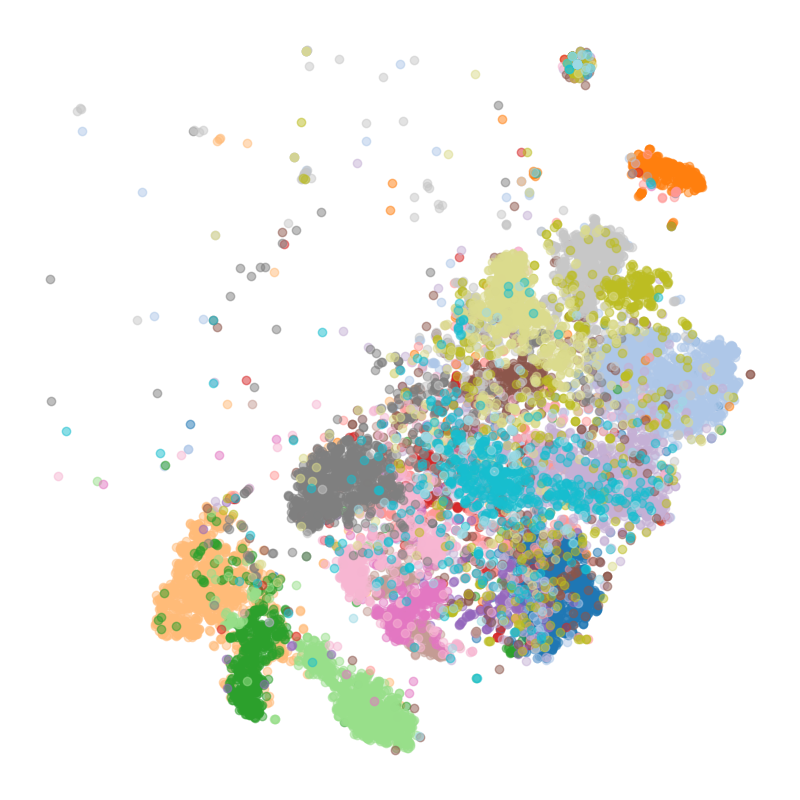}
\\
(a) PFENet & (b) SS-PFENet & (c) PFENet & (d) SS-PFENet
\end{tabular}
\caption{Visual comparison between feature spaces from PFENet trained with and without the purposed self-supervision. (a) and (b) were trained on PASCAL-5i~\cite{shaban2017one} excluding \emph{fold2} classes, (c) and (d) were trained on COCO excluding \emph{fold3} classes.}
\label{fig:feature_space}
\end{figure}

To further understand the impact of the proposed self-supervised task, we used t-SNE~\cite{ljpvd2008visualizing} to visualise the foreground embedding from query images in Figure~\ref{fig:feature_space}. Each dot corresponds to the global average pooled query foreground feature from a testing episode, and different colours represent different classes. The point clouds of different classes tend to be better disentangled, which qualitatively supports our theory and explains the observed results.

\subsection{Ablation Studies}
\label{sec:ablation}

\subsubsection{Model Architecture}
\label{sec:ablation-model-arch}

To test the applicability of the proposed self-supervision across different architectures, further experiments were performed on PANet~\cite{wang2019panet}, an intrinsically different architecture that adopts cosine similarity to compare both foreground and background prototypes with each pixel and make predictions. 
As shown in Table~\ref{tab:pascal_table} and Table~\ref{tab:coco_table}, the proposed method achieved $4.1\%$ and $2.6\%$ absolute improvement on PASCAL-5i 1-shot and 5-shot tasks as well as $5.2\%$ and $8.0\%$ absolute improvement on COCO 1-shot and 5-shot tasks.
The consistent improvement on multi-class query images was also observed across architectures as shown in Figure~\ref{fig:num_class}(b).

\subsubsection{Pseudo-class Generation}
We compared the following three other different pseudo-class generation processes: 1) Gridding: divide the image into $10\times 10$ grids evenly, 2) SLIC~\cite{achanta2012slic}: segment image through iterative clustering, setting compactness to 10 and n\_segments to 100, 3) HED contour detector~\cite{xie2015holistically}: leverage deep learning supervision to detect edges. The meanIoU on the PASCAL-5i 1-shot task are reported in Table~\ref{tab:superpixel_table}.
Gridding leads to a significant drop of performance to 57.5\%, lower than being without self-supervision.
The possible reason may be the presence of multiple classes together with the background in the same grid, making the pseudo-classes misleading.
We chose to use the second-best algorithm Felzenszwalb instead of HED because HED was pre-trained on the SDS dataset~\cite{hariharan2014simultaneous}, part of our test dataset, which may cause potential information leakage.
\newcommand{\superpixeltable}{\small

\begin{tabular}{|c|c|cccc|}
\hline
method  & PFENet & \multicolumn{4}{|c|}{SS-PFENet} \\
        &        & gridding & SLIC & HED           & Felzenszwalb \\\hline
meanIoU & 60.7   & 57.5     & 62.9 & \textbf{63.3} & 63.2 \\
\hline
\end{tabular}}
\begin{table}[ht]
    \centering
    \small

\begin{tabular}{|c|c|cccc|}
\hline
method  & PFENet & \multicolumn{4}{|c|}{SS-PFENet} \\
        &        & gridding & SLIC & HED           & Felzenszwalb \\\hline
meanIoU & 60.7   & 57.5     & 62.9 & \textbf{63.3} & 63.2 \\
\hline
\end{tabular}
    \caption{Performance of ResNet-50 PFENet on PASCAL-5i 1-shot task with self-supervision using different superpixel segmentation algorithms.}
    \label{tab:superpixel_table}
\end{table}

We also compared three different pseudo-class sampling strategies: 1) select the superpixel with the highest activation score calculated in Eq.~\ref{eq:activation_score}, 2) sample a superpixel with the top-5 activation scores, 3) sample a superpixel from the entire image randomly. The meanIoUs on the PASCAL-5i 1-shot task are reported in Table~\ref{tab:toppick_table}, with the "top-5" strategy performing the best. This is understandable as on the one hand, top-5 provides more pseudo-classes than top-1 for training; on the other hand, selecting a random superpixel without ranking may result in over-segmenting, for example, a background region, e.g. sky, to be different to other background regions, e.g. other superpixels of the sky, thus misguiding the model.

\newcommand{\toppicktable}{\small
\begin{tabular}{|c|c|ccc|}
\hline
method                       & PFENet & \multicolumn{3}{|c|}{SS-PFENet} \\
pesudo-class choice range    &        & all  & top-5         & top-1 \\\hline
meanIoU                      & 60.7   & 58.5 & \textbf{63.2} & 63.0  \\
\hline
\end{tabular}}
\begin{table}[ht]
    \centering
    \small
\begin{tabular}{|c|c|ccc|}
\hline
method                       & PFENet & \multicolumn{3}{|c|}{SS-PFENet} \\
pesudo-class choice range    &        & all  & top-5         & top-1 \\\hline
meanIoU                      & 60.7   & 58.5 & \textbf{63.2} & 63.0  \\
\hline
\end{tabular}
    \caption{Performance of ResNet50-PFENet on PASCAL-5i task with self-supervision when choosing pseudo-class from superpixels with the top-n activation scores.}
    \label{tab:toppick_table}
\end{table}

\textcolor{black}{
\subsubsection{Effect of Novel Class in Training Background}
Though there is no label of testing classes available during training, their existences in the training images could still lead to advantage when the sampled \emph{pseudo-classes} overlap with those regions. This ablation study examines the performance of the purposed method without such advantage by masking all pixels belonging to the testing classes in training images so these testing class objects will not be part of any pseudo-classes during training. As shown in Table~\ref{tab:mask_novel_table}, the meanIoU decreased only by $0.1\%$ and $0.4\%$ respectively for PANet and PFENet architecture. This minor decrease indicates the ability of the proposed self-supervision to generalise to completely unseen novel classes.
}

\newcommand{\masknoveltable}{\begin{tabular}{|c|cccc|c|}
\hline
method                      & fold1         & fold2         & fold3         & fold4         & mean \\
\hline
PANet\cite{tian2020prior}   & 42.3          & 58.0          & 51.1          & 41.3          & 48.2 \\ 
SS-PANet                    & \textbf{49.3} & \textbf{60.8} & 53.9          & \textbf{45.2} & \textbf{52.3} \\
SS-PANet-masked             & 49.1          & 60.7          & \textbf{54.0} & 45.0          & 52.2 \\
\hline
PFENet\cite{tian2020prior}  & \textbf{61.7} & 69.5          & 55.4          & 56.3          & 60.7 \\
SS-PFENet                   & 58.9          & \textbf{69.9} & \textbf{66.4} & \textbf{57.7} & \textbf{63.2} \\
SS-PFENet-masked            & 58.8          & 69.7          & 65.4          & 57.3          & 62.8 \\
\hline
\end{tabular}}
\begin{table}[ht]
    \centering
    \begin{tabular}{|c|cccc|c|}
\hline
method                      & fold1         & fold2         & fold3         & fold4         & mean \\
\hline
PANet\cite{tian2020prior}   & 42.3          & 58.0          & 51.1          & 41.3          & 48.2 \\ 
SS-PANet                    & \textbf{49.3} & \textbf{60.8} & 53.9          & \textbf{45.2} & \textbf{52.3} \\
SS-PANet-masked             & 49.1          & 60.7          & \textbf{54.0} & 45.0          & 52.2 \\
\hline
PFENet\cite{tian2020prior}  & \textbf{61.7} & 69.5          & 55.4          & 56.3          & 60.7 \\
SS-PFENet                   & 58.9          & \textbf{69.9} & \textbf{66.4} & \textbf{57.7} & \textbf{63.2} \\
SS-PFENet-masked            & 58.8          & 69.7          & 65.4          & 57.3          & 62.8 \\
\hline
\end{tabular}
    \caption{meanIoU results on PASCAL-5i~\cite{shaban2017one}. All models adopts ResNet-50 as backbone and best performance achieved by same model are bolded. ``SS-'' refers to models with self-supervision and ``-mask'' refers to models trained with pixels belonging to novel classes masked.}
    \label{tab:mask_novel_table}
\end{table}

\subsubsection{Effect of $\alpha$ Parameter}

\newcommand{\alphatable}{\begin{tabular}{|c|ccccc|}
\hline
$\alpha$    & 0.0   & 0.25  & 0.50          & 0.75  & 1.0   \\\hline
meanIoU     & 60.7  & 62.5  & \textbf{63.2} & 61.2  & 59.5  \\\hline
\end{tabular}}
\begin{table}[ht]
    \centering
    \begin{tabular}{|c|ccccc|}
\hline
$\alpha$    & 0.0   & 0.25  & 0.50          & 0.75  & 1.0   \\\hline
meanIoU     & 60.7  & 62.5  & \textbf{63.2} & 61.2  & 59.5  \\\hline
\end{tabular}
    \caption{Performance of ResNet-50 PFENet on PASCAL-5i~\cite{shaban2017one} with self-supervision using different $\alpha$ values.}
    \label{tab:alpha_table}
\end{table}

\textcolor{black}{
We varied the self-supervised loss scaling factor $\alpha$ and report the performance achieved on PASCAL-5i~\cite{shaban2017one} in Table~\ref{tab:alpha_table}. $\alpha = 0.5$ appears to be the optimal value. This is reasonable because the pseudo labels generated by superpixel segmentation are inaccurate (over-segmentation) and thus less aligned with our objective than ground-truth annotations.
}

\section{Conclusion}
\label{sec:conclusion}
In this work, we first raised the issue of the dependency between the few-shot segmentation performance and the inherent complexion of individual \emph{query} images, specifically, the presence of additional classes. We have provided quantitative evidence showing accuracy decreases in segmentation for \emph{novel} classes, as a result of more latent objects appearing in the background of the \emph{target} class of interest.
We directly address the lack of background class supervision, by proposing a novel strategy that generates self-supervising pseudo-labels for the non-target classes. A set of extensive experimental results have shown that the proposed approach consistently improved the few-shot segmentation performance on images that contain multiple classes of objects, for different network architectures on multiple datasets.
Finally, we showed that different classes were indeed better disentangled in the embedding space using our method.
The proposed methodology is not limited by the choice of pseudo-supervision generation method, here we proposed a superpixel-based approach that is simple and effective. 
Future work will investigate alternatives that, for instance, base on higher-level features to scale better to intra-class variations such as shape and colour variations, and their potentially latent relationship to the model adaptability and predictivity. 

\newpage
{\small
\bibliography{mybib}
}

\newpage
\section{Supplementary Material}
\begin{figure}[ht]
    \centering
    \begin{tabular}{c}
    \includegraphics[width=0.5\linewidth]{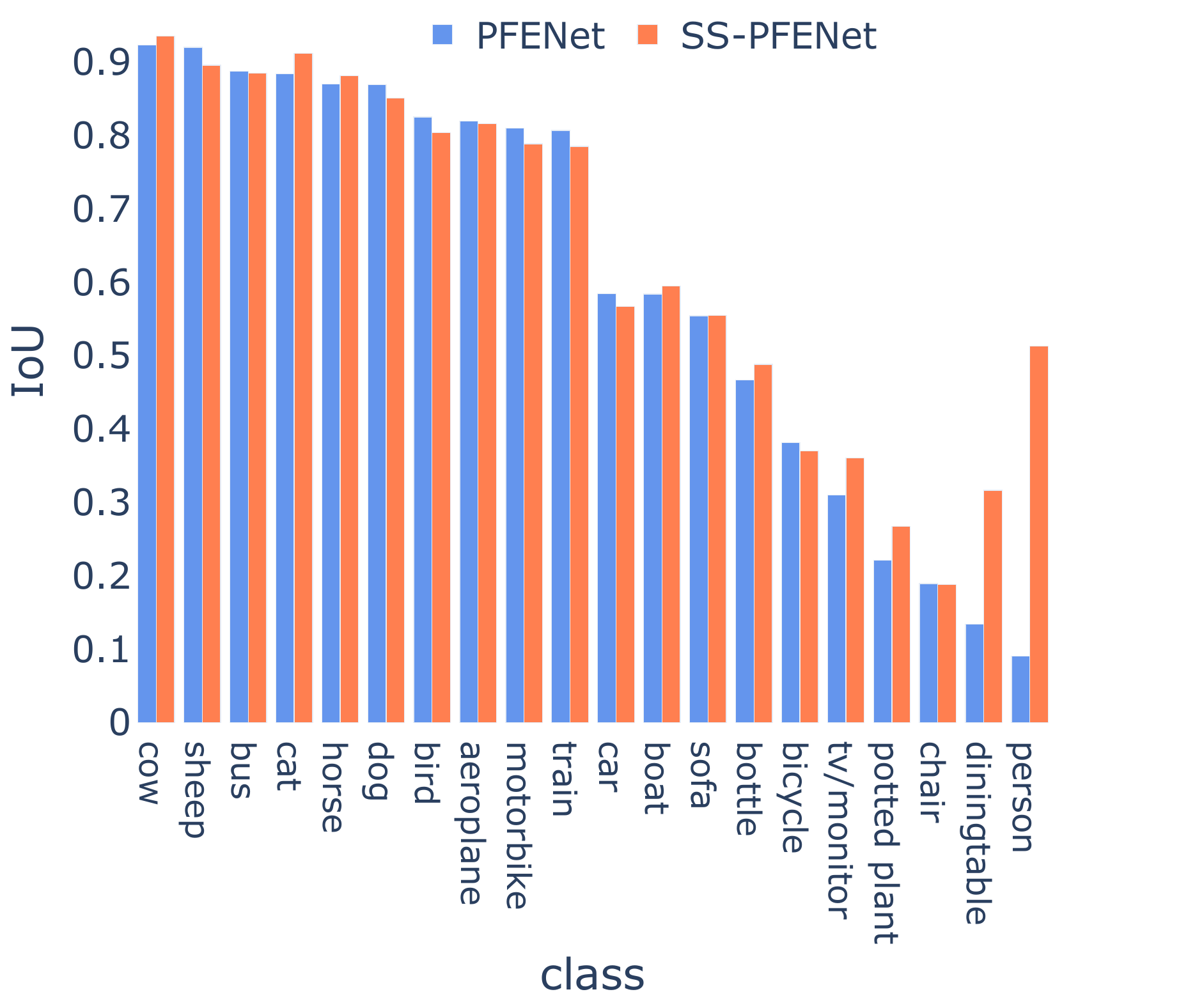} \\
    (a) PASCAL-5i\\
    \includegraphics[width=1.0\linewidth]{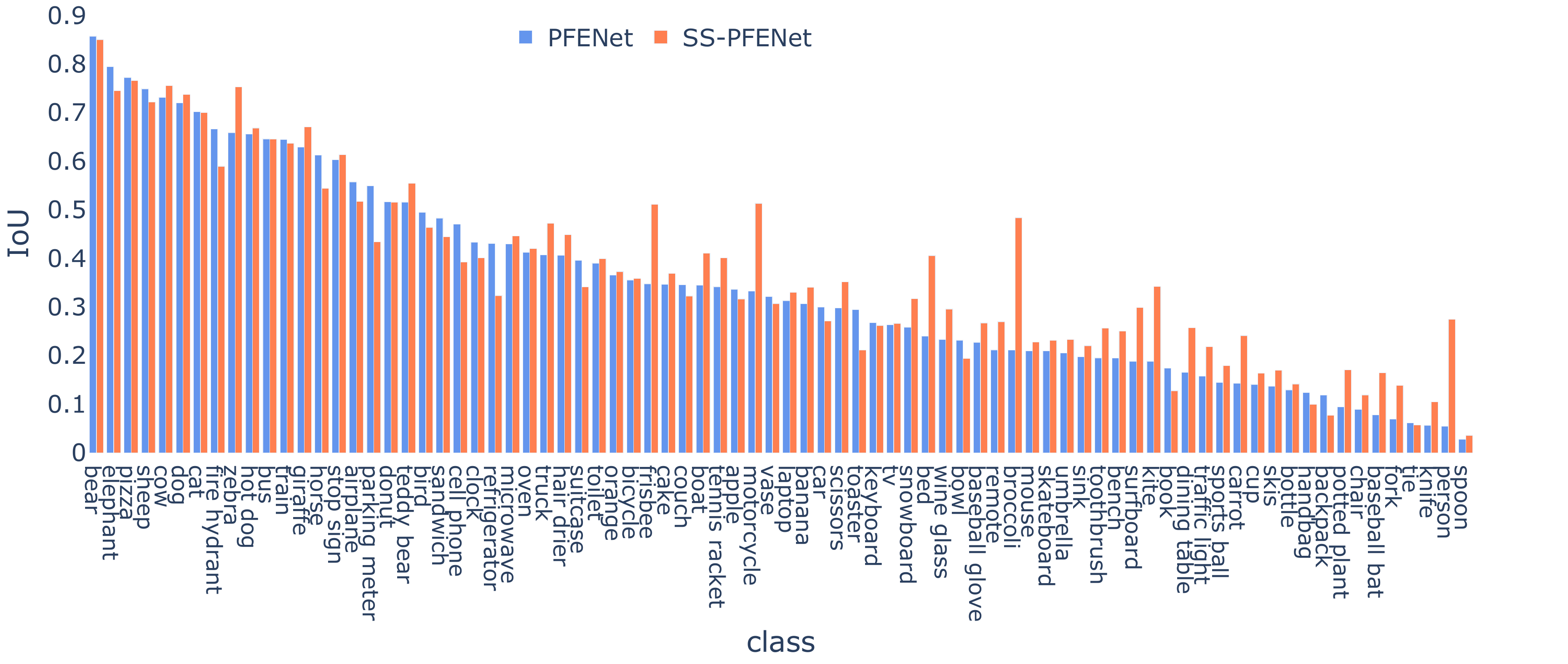} \\
    (b) COCO-20i\\
    \end{tabular}
    \caption{Class IoUs reached by PFENet with and without self-supervision on (a) PASCAL-5i and (b) COCO-20i dataset. X-axis corresponds to different classes in descending order of performance reached by PFENet without self-supervision.}
    \label{fig:my_label}
\end{figure}

\begin{figure}[ht]
    \centering
    \includegraphics[width=0.5\linewidth]{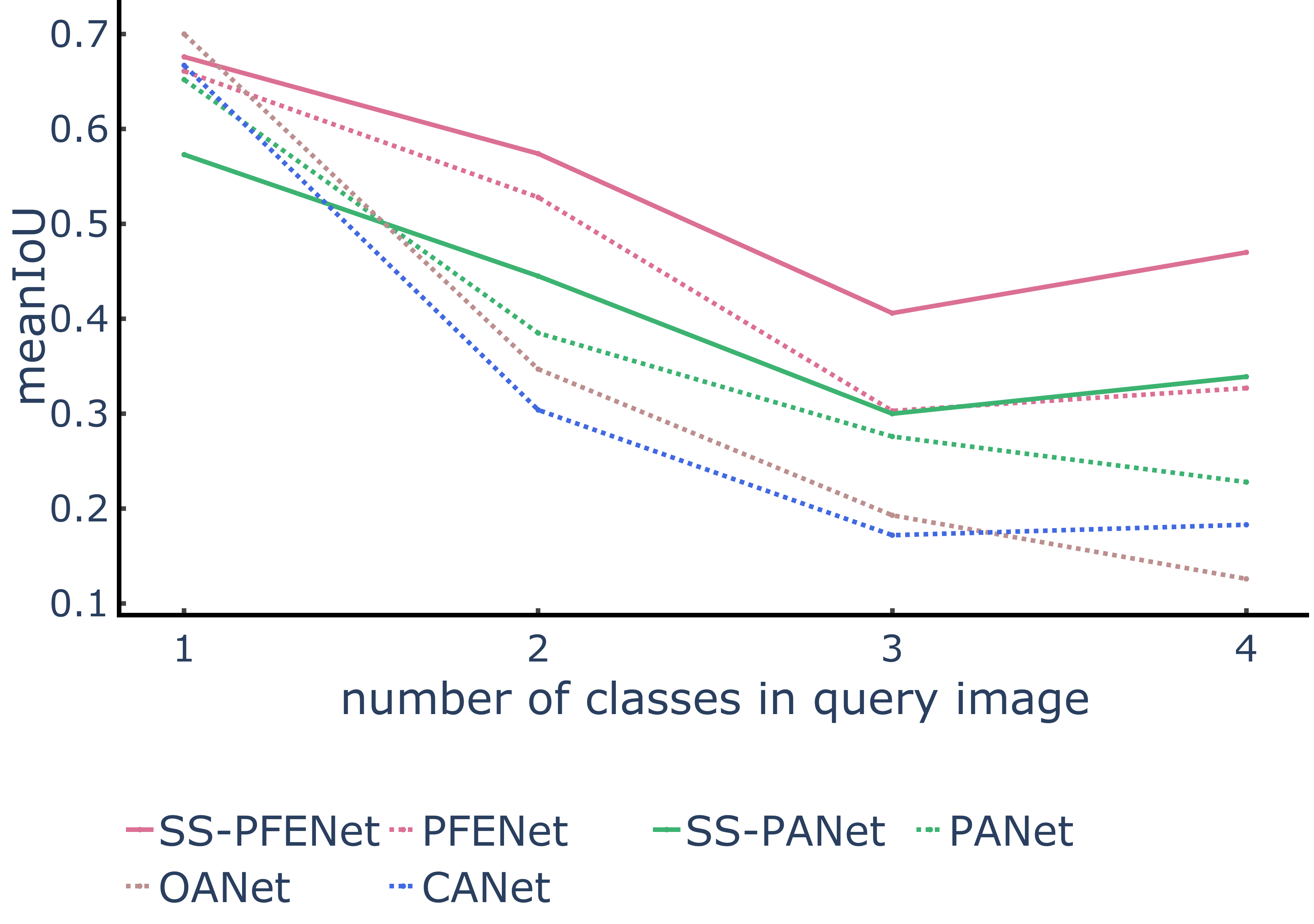}
    \caption{Performance of various few-shot semantic segmentation methods versus the number of semantic classes in the query image.
    Different colours represent different methods and the prefix “SS-” denotes models with the proposed self-supervision.
    Existing methods tend to perform similarly well when there is a single class in the query, but their performance tends to decrease when more classes are present.}
\end{figure}

\begin{figure}[ht]
    \centering
    \includegraphics[width=0.9\linewidth]{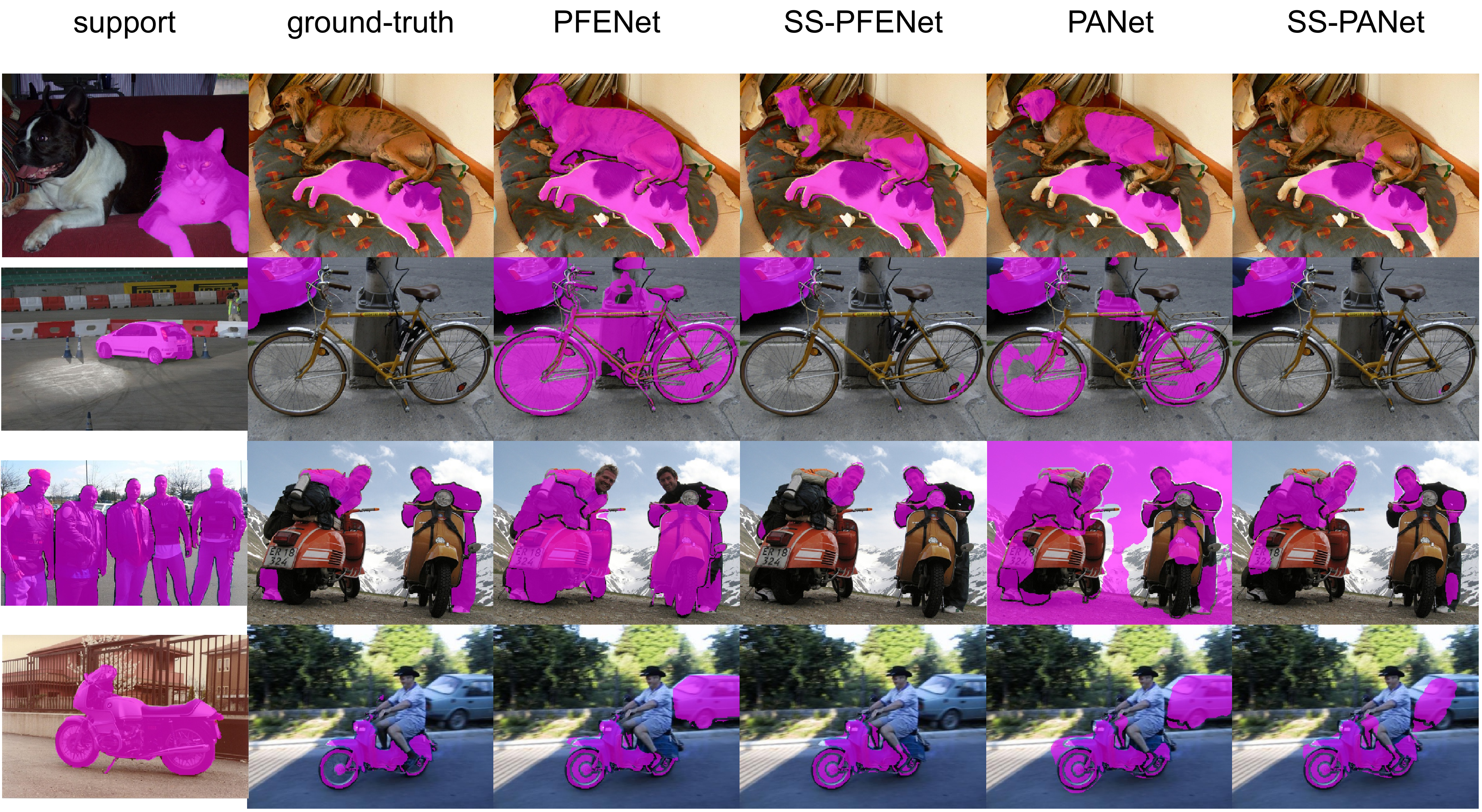}
    \caption{Qualitative results achieved by PFENet and PANet with and w/o the purposed self-supervision.}
\end{figure}

\end{document}